# Generating Navigable Semantic Maps
# from Social Sciences Corpora

Thierry Poibeau & Pablo Ruiz

Laboratoire LATTICE
CNRS & Ecole normale supérieure & Université Paris 3 Sorbonne Nouvelle
1 rue Maurice Arnoux, 92120 Montrouge – France

{thierry.poibeau,pablo.ruiz.fabo}@ens.fr

It is now commonplace to observe that we are facing a deluge of online information. Researchers have of course long acknowledged the potential value of this information since digital traces make it possible to directly observe, describe and analyze social facts, and above all the co-evolution of ideas and communities over time.

However, most online information is expressed through text, which means it is not directly usable by machines, since computers require structured, organized and typed information in order to be able to manipulate it. Our goal is thus twofold:

1. Provide new natural language processing techniques aiming at automatically extracting relevant information from texts, especially in the context of social sciences, and connect these pieces of information so as to obtain relevant socio-semantic networks;
2. Provide new ways of exploring these socio-semantic networks, thanks to tools allowing one to dynamically navigate these networks, de-construct and re-construct them interactively, from different points of view following the needs expressed by domain experts.

Nodes in a network can be a multitude of different entities (people, keywords or more complex notions such as ideas or opinions) as long these entities can be formalized. The content of the nodes as well as the structure of the network must be specifiable by the expert depending on their goal, framework or theoretical background.

The approach is thus not linked to any particular theory. Instead we want to promote a practical approach making it easy to manipulate and exploit the content of texts so as to dynamically construct meaningful representations and maps. This approach makes it possible to observe regular patterns and behaviors without having to define a priori categories and groups (such as individuals *vs* society), which is why we can say that the approach is not linked to any specific theory. Instead, we think it can be more rewarding and more interesting to observe aggregates (i.e. specific groups of nodes in a network), and their stability and instability over time (Latour *et al.*, 2012).

In order to illustrate our approach, we now turn to a discussion of an example of our recent work. This work involved computational linguists as well as domain experts.

## 1   A Practical Example: Mapping the 2007-2008 Financial Crisis

The 2007-2008 financial crisis was a dramatically complex event and the political responses to this event were at least as complex. These responses can be studied thanks to a huge amount of documents produced by various bodies during the crisis and made available since then.

An American initiative aims at studying the response of the American authorities to the crisis through PoliInformatics, defined as "an interdisciplinary field that promotes diverse methodological approaches to the study of politics and government" (http://poliinformatics.org/). We participated in the first PoliInformatics challenge, as we describe in following.

The organizers of the challenge made available a series of documents on the 2007-2008 financial crisis. The shared task consisted in developing solutions to address questions such as "Who was the financial crisis?" or "What was the financial crisis?". Of course, these questions are too complex to receive a simple and direct answer. So our strategy has been to provide tools to process and visualize the most relevant data, so that experts can easily navigate into this flow of information and make sense of the data. While we believe in semi-automatic corpus exploration, we do not think it is possible or even desirable to provide fully automatic answers to the above questions.

We have as far as possible used available tools to extract and visualize information. More precisely, we have used the Stanford Named Entity Recognizer (Finkel et al., 2005) and the Cortext platform (http://www.cortext.net/) for information extraction. As for data visualization, we have used Gephi (Bastian et al., 2009) to observe semantic and social networks, and the Cortext platform to observe the evolution of the domain over time. However, these tools are not enough to obtain meaningful representations: for example, new developments are necessary for named entity normalization and linking, esp. to link text with ontologies (Ruiz and Poibeau, 2015). The result should then be filtered following precise, domain-dependent criteria, so as to obtain navigable and readable maps with the most salient information.

## 2   Technical Overview

**Named Entity Recognition and Normalization**

The first step was to extract named entities from the different corpora. Named Entity Recognition is a mature technology that has been used in several Digital Humanities projects (see Van Hooland et al., 2013 for a discussion of some recent projects). However, the use of NER in the analysis of political science texts seems to have been limited, e.g. Grimmer and Stewart's (2013) survey of text analytics for political science includes no discussion of this technology.

In order to perform entity extraction, we used the Stanford NER, based on Conditional Random Fields, with MUC tags (Time, Location, Organization, Person, Money, Percent, Date) (Finkel et al., 2005). Certain entities appear under different forms. For instance, "Standard and Poor" might occur as "Standard & Poor", "S&P" or "Standard & Poor's executive board" (this last sequence in fact refers to a slightly different named entity); in a similar fashion, a person as "Mary Schapiro" may appear as "Schapiro", or "Miss Schapiro" or "Chairman Schapiro". We implemented a simple normalization method based on the maximization of common sub-sequence between two strings and obtained qualitatively good results when compared to other more sophisticated algorithms (Gottipati and Jiang, 2011; Rao et al., 2011). Entity linking could then be applied on the result.

## Visualizing entities

We used the Gephi software (Bastian et al., 2009) so as to create graphs for each corpus, such that:

- a node corresponds to a cluster of persons or organizations in the corresponding corpus;
- an edge between two nodes corresponds to the number of co-occurrences of the two nodes within the same sentence in the corpus.

We chose to consider persons and organizations together since they can play a similar role in the event, and metonymy is often used, so that a person can refer to a company (and vice versa).

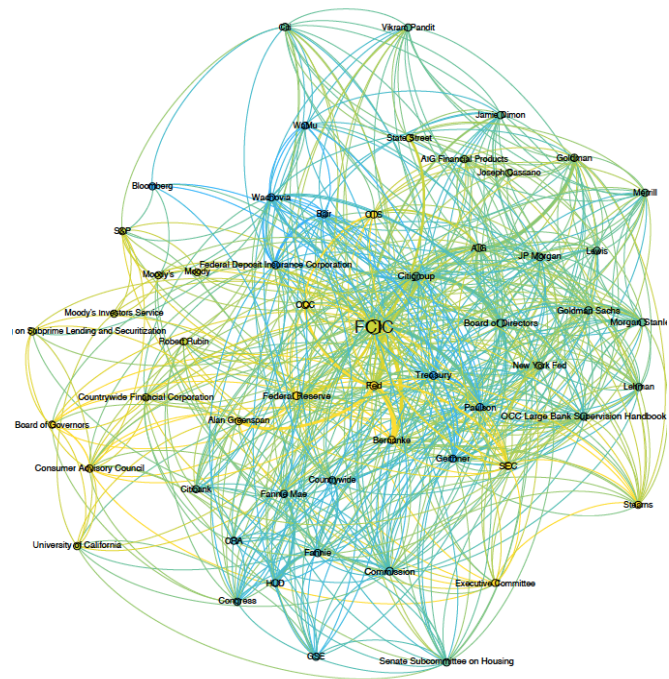

**Figure 1:** Visualization of links between entities with Gephi

Some results can be seen on figure 1 (which is hardly readable in small format but can be interactively explored by the experts in the field on a computer). Some links correspond to well establish relations like the link between an organization and its CEO (see for ex. the link between Scott Polakoff and OTS, or between Fabrice Tourre and Goldman Sachs). However, we are also able to extract less predictable links that could be of interest for scholars and experts in the field. As an example, we observe a link between the Fed Consumer Advisory Council and the Board of Governors (for ex. Bernanke, Mark Olson, and Kevin Warsh) since the first group of people (the council) warns vigorously the Board of Governors about the crisis. An interesting methodological issue to consider when elaborating these networks is that different automatic linguistic analyses can affect the complexity of the network: For instance, depending on the strategy to calculate intra-document coreference (e.g. whether we consider pronouns referring to an entity as an instance of the entity or not), the amount of edges in the graph will vary. Rieder and Röhle (2012) have discussed interpretation problems for visualizations, and Rieder (2010) has commented on how different graph layout algorithms can lead to representations that promote opposite interpretations of a network for the same corpus. We are as well interested in exploring how the computational linguistics tools employed in order to assess co-occurrences, even before applying a graph layout, can influence the graphs ultimately produced.

**Visualizing temporal evolution**

The visualizations we produced should be explored and tested by specialists who could evaluate their real benefits. A historic view on the data would also be useful to analyze the dynamics and the evolution of the crisis, through for example the evolution of terms associated with named entities over different periods of time.

We tried to explore and visualize the temporal evolution of the financial crisis, more specifically the evolution of the perceived role of organizations over time. To do so, we produced Sankey diagrams of the correlation of organizations and domain related terms in the corpus. With this strategy, Sankey diagrams take into account the temporal evolutions of entities and actions along the crisis.

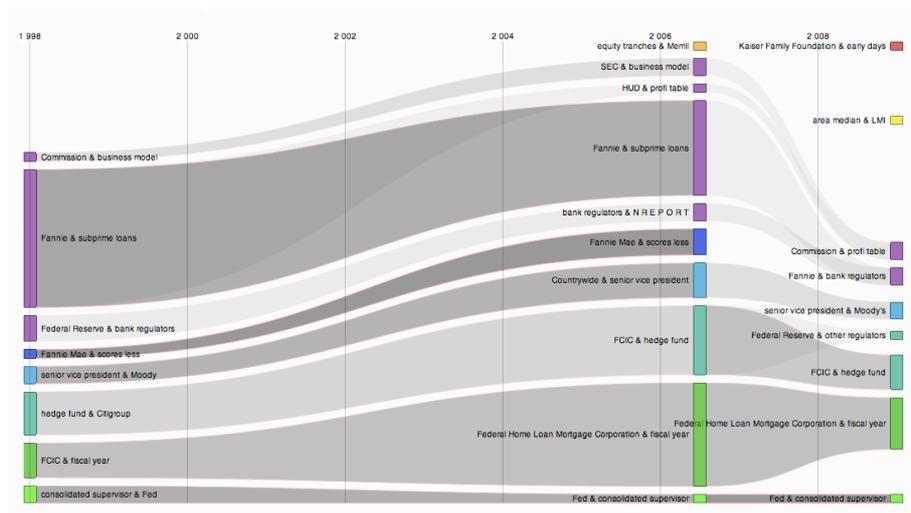

**Figure 2:** Evolution of the links between named entities and topics over time

Figure 2 reveals a modification in the data between 2006 and 2008, a period which approximates the start of the financial crisis. For instance, the stream in purple in this graph reveals many co-occurrences of Fannie Mae and subprime loans for the period 1990-2007 while for the period 2008-2010, Fannie Mae is more closely associated with 'bank regulators', or 'Federal Financial Services Supervisory Authority'. In a more general way, all the streams of data represented in the diagram are dramatically modified after 2007.

## 3   Evaluation and Future Work

The work presented here has been evaluated by a panel of experts in the field. They assessed the utility of the tools and helped us define ways to improve these first results. Perspectives are thus twofold: on the one hand enhance data analysis so as to provide more relevant maps and representations, and on the second hand work closely with domain experts and provide interactive ways of navigating the data. Concerning interactions with experts, it is clear that end users could provide a very valuable contribution in the selection of relevant data as well as in the way they are linked and mapped. Some experiments are currently being done with a focus group gathering social science as well as information science experts. They will assess that the solution is useful and workable and more importantly, will give feedback so as to provide better solutions.

# 4 Acknowledgements

This work has received support of Paris Sciences et Lettres (program "Investissements d'avenir" ANR-10-IDEX-0001-02 PSL*) and of the laboratoire d'excellence TransferS (ANR-10-LABX-0099). Pablo Ruiz is funded thanks to a grant from the Region Ile-de-France.